\documentclass{article}

\PassOptionsToPackage{numbers, compress}{natbib}

\usepackage{amsmath}
\usepackage{algorithm}
\usepackage{algpseudocode}
\usepackage{graphicx} 
\usepackage{titletoc}

 \usepackage[preprint]{neurips_2026}

\usepackage[utf8]{inputenc} %
\usepackage[T1]{fontenc}    %
\usepackage{hyperref}       %
\usepackage{bookmark}       %
\usepackage{url}            %
\usepackage{booktabs}       %
\usepackage{amsfonts}       %
\usepackage{nicefrac}       %
\usepackage{microtype}      %
\usepackage[table]{xcolor}  %
\usepackage{colortbl}       %
\usepackage{multirow}       %
\usepackage{makecell}       %
\usepackage{array}          %
\usepackage{tabularx}       %
\usepackage{amssymb}        %
\usepackage{bm}             %
\usepackage{enumitem}       %
\usepackage{siunitx}        %

\definecolor{ourhighlight}{RGB}{214,232,250}
\definecolor{lightblue}{RGB}{222,239,255}
\definecolor{verylightgray}{gray}{0.93}
\definecolor{ourhighlight}{RGB}{220, 245, 220}
\definecolor{figGreen}{HTML}{4FA483}  
\definecolor{figBlue}{HTML}{3A8FCC}    
\definecolor{figPurple}{HTML}{7E5FE6}   
\definecolor{figOrange}{HTML}{D67A1A} 
\definecolor{verylightpurple}{HTML}{F4EFFF}

\newcommand{\method}{VideoMLA}

\title{VideoMLA: Low-Rank Latent KV Cache for Minute-Scale Autoregressive Video Diffusion}

\author{
\textbf{Hidir Yesiltepe}$^1$ \quad
\textbf{Jiazhen Hu}$^1$ \quad
\textbf{Tuna Han Salih Meral}$^1$ \quad
\textbf{Adil Kaan Akan}$^2$ \\
\textbf{Kaan Oktay}$^2$ \quad
\textbf{Hoda Eldardiry}$^1$ \quad
\textbf{Pinar Yanardag}$^1$ \\
$^1$Virginia Tech \qquad $^2$fal \\ [0.5em]
\normalsize{Project Page: \url{https://videomla.github.io}}}

\begin{document}

\maketitle

\begin{abstract}
Long-rollout causal video diffusion has converged on a fixed-size sliding-window KV cache, with recent progress innovating \emph{within} this layout by changing which tokens occupy the window or how their positions are encoded. The per-head KV layout itself, a dominant contributor to streaming memory and latency, has been mostly left unchanged. In this paper, we present the first study of Multi-Head Latent Attention (MLA) in video diffusion. \method{} replaces per-head keys and values with a shared low-rank content latent and a shared decoupled 3D-RoPE positional key, reducing per-token KV memory by $92.7\%$ at every cached layer. We further investigate why MLA succeeds in video diffusion even though the spectral assumption often used to motivate it in language models does not hold: pretrained video attention is not low-rank, with 99\%-energy effective rank far above any practical latent dimension. \method{} retains quality at compression ratios where direct spectral approximation would predict large reconstruction error. We show that the MLA bottleneck, rather than the pretrained spectrum, determines the effective rank: both spectral and random initialization occupy nearly the full rank budget from initialization, and training preserves this budget while adapting within it. On VBench, \method{} matches short-horizon streaming video diffusion baselines, achieves the best overall score at long horizons among evaluated methods, and improves throughput by $1.23\times$ on a single B200.
\end{abstract}

\section{Introduction}
\label{sec:intro}
Causal video diffusion models \cite{huang2025self, cui2025self, liu2025rolling, zhu2026causal, yi2025deep, yesiltepe2025infinity, lu2025reward, yin2025slow, yang2025longlive, chen2025sana, yang2026anchor, kim2026memrope, zhao2026relax, li2026rolling} have gained traction as the dominant approach to streaming, long-horizon video generation. Distilled from bidirectional teachers, they generate frames \cite{zhu2026causal, deng2024autoregressive, jin2024pyramidal, henschel2025streamingt2v} or chunks autoregressively while attending to a rolling key-value (KV) cache of past frames, producing minute-long videos at interactive rates on a single GPU. As models scale toward longer rollouts, the per-head KV cache increasingly defines the operating point.  At Wan-1.3B scale~\cite{wan2025wan}, each cached token stores $2 \times 12 \times 128 = 3{,}072$ dense KV scalars per layer, accounting for keys and values across 12 heads with 128 channels each. With a 21-latent-frame cache, 1,560 tokens per latent frame, and 30 transformer layers, the dense KV cache contains 3.02B scalars, or about 6.0GB in bf16/fp16. This footprint explains why recent streaming systems use fixed-size sliding-window caches: retaining all past KV states would grow linearly with rollout length. However, fixing the window only bounds the number of cached tokens; it does not reduce the per-token, per-layer cost of the per-head KV layout. Reducing this layout is therefore a direct lever for longer horizons, larger batches, and faster inference.

The dominant line of recent work treats the cache as a fixed-size sliding window and innovates inside it. CausVid~\cite{yin2025slow} initiated this thread by converting bidirectional diffusion into causal autoregressive generation via distribution matching distillation, with a sliding KV cache from inception. Self-Forcing~\cite{huang2025self} closed the train--test gap by conditioning training on self-generated frames within the same rolling cache. Subsequent work refined this recipe through attention-sink, token-selection, and compressed-memory mechanisms for long-range consistency~\cite{liu2025rolling,yi2025deep,lu2025reward,yu2025videossm,kim2026memrope, zhang2025packing}, training strategies for multi-minute rollouts and prompt switching~\cite{cui2025self,yang2025longlive, gao2025longvie, yesiltepe2025infinity}, improved distillation objectives~\cite{zhu2026causal,lu2025reward}, and positional reparameterization such as Infinity-RoPE~\cite{yesiltepe2025infinity}. However, these methods all preserve the per-head KV layout that fills the window in the first place: they redistribute, reweight, compress over time, or reposition cached tokens without reducing the per-token KV state.

A second, complementary line changes the attention computation itself. SANA-Video~\cite{chen2025sana} replaces softmax attention with block-causal linear attention, removing the conventional KV cache and using a constant-memory cumulative state for long-video generation. SCD~\cite{bai2026causality} reduces cached state by routing temporal reasoning through a 25-layer causal encoder and using a 10-layer frame-wise decoder, so only the encoder layers cache. Under the same Wan cache geometry, this reduces dense KV storage by $16.7\%$. \method{} is orthogonal: it keeps all 30 self-attention layers cached but reduces each token's cached state from 3072 to 224 scalars, yielding an $11.4\times$ smaller cache than SCD for the same 21-latent-frame window. Thus, rather than changing which tokens are cached, how they are positioned, or how many layers cache, \method{} targets the remaining factor directly: the per-token KV layout at every cached self-attention layer.

\begin{figure}[t]
    \centering
        \includegraphics[width=1.0\linewidth]{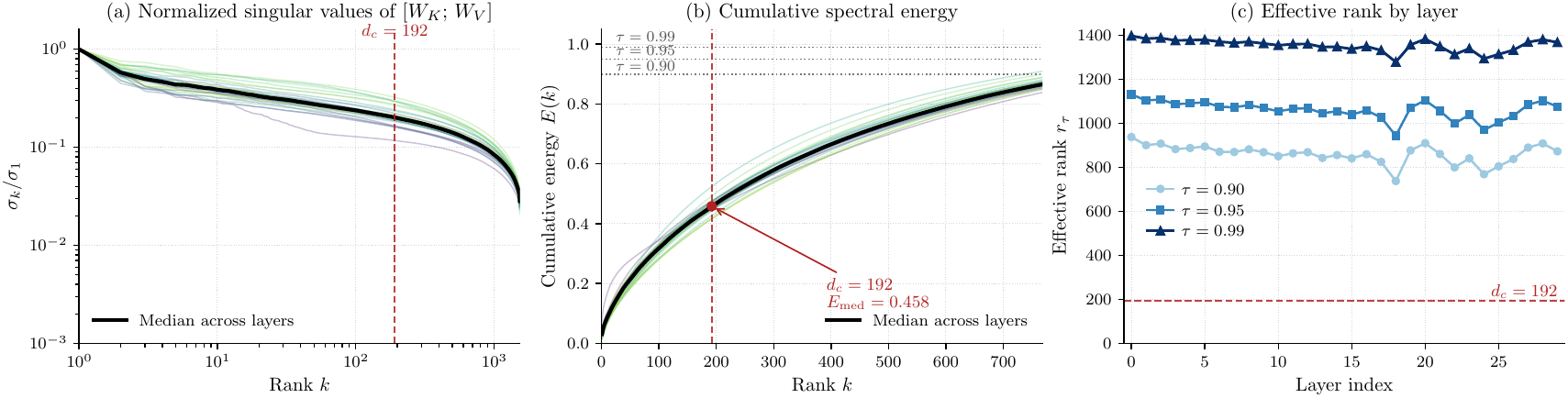} \\
    \caption{\textbf{Pretrained video diffusion attention is not low-rank, unlike in language models.}
Singular value analysis of $[W_K;\, W_V] \in \mathbb{R}^{3072 \times 1536}$ across the 30 transformer blocks of Wan2.1-T2V-1.3B. At $d_c = 192$, the median layer captures only $E_{\mathrm{med}} = 0.458$ of the spectral energy, and the 99\%-energy effective rank exceeds 1300 in every layer.}
\label{fig:svd_spectrum}
\end{figure}

\begin{figure}[t]
    \centering
        \includegraphics[width=1.0\linewidth]{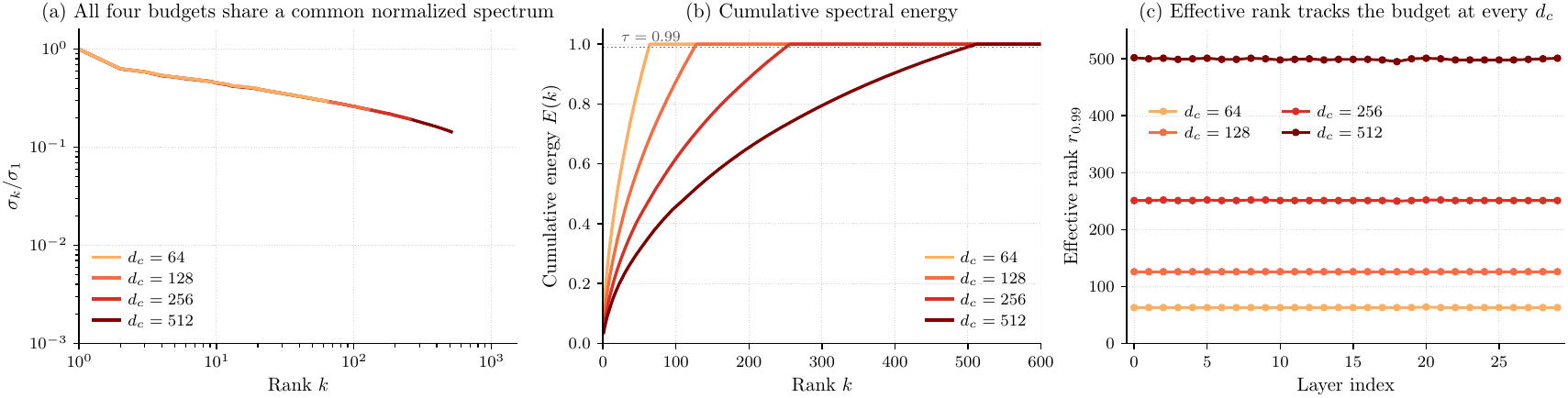} \\
    \caption{\textbf{The composed operator occupies its full rank-$d_c$ budget at every $d_c$ and every layer.}
Singular value analysis of the composed operator $M_\text{learned} = [W^K_\uparrow W^{KV}_\downarrow; W^V_\uparrow W^{KV}_\downarrow]$ for SVD-initialized \method{} students at $d_c \in \{64, 128, 256, 512\}$. \textbf{(a)} Median normalized spectra share a common envelope, truncated at $d_c$. \textbf{(b)} Cumulative spectral energy. \textbf{(c)} Layer-wise 99\%-energy effective rank: $r_{0.99} \approx 0.98\,d_c$ at every budget, uniformly across depth. The composed operator's rank is determined by the architectural bottleneck, not by the spectral structure of the dense source.}
\label{fig:learned_rank}
\end{figure}

In this paper, we intervene on the per-head layout itself. Building on Multi-Head Latent Attention (MLA)~\cite{liu2024deepseek}, we present \method{}, the first MLA-style latent KV cache for autoregressive video diffusion. \method{} replaces per-head keys and values with a shared low-rank content latent and a head-shared decoupled 3D-RoPE positional key, reducing per-token KV memory by $92.7\%$ at \emph{every} cached layer. This raises a puzzle: MLA is usually motivated by low-rank pretrained $W_K,W_V$~\cite{liu2024deepseek, ji2025towards}, yet Wan-1.3B (Fig.~\ref{fig:svd_spectrum}) has 99\%-energy rank far above practical latent dimensions. \method{} nonetheless retains quality where direct spectral approximation would incur large reconstruction error (Fig.~\ref{fig:learned_rank}). We show that the MLA bottleneck, not the pretrained spectrum, determines the effective rank: SVD and random initialization both nearly saturate the rank budget, which training preserves with little spectral change. The design question therefore shifts from \emph{what is the intrinsic rank?} to \emph{what latent budget preserves video quality?} Our contributions are summarized as follows:

\begin{itemize}
    \item \textbf{Latent KV caching for video diffusion.}
    We introduce \method{}, an MLA-style autoregressive video diffusion model that replaces per-head keys and values with a shared content latent and a head-shared decoupled 3D-RoPE key, reducing per-token KV memory by $92.7\%$ at every cached layer.

    \item \textbf{A spectral puzzle and rank-budgeted resolution.}
    We show that Wan-1.3B video attention is not low-rank: the 99\%-energy effective rank of $[W_K; W_V]$ far exceeds practical latent dimensions. \method{} nevertheless retains quality, while both SVD and random initialization saturate the imposed rank budget from initialization and preserve it during training.

    \item \textbf{Efficient long-horizon generation.}
    We identify the NoPE/RoPE allocation that preserves visual fidelity and motion consistency at minute-scale horizons. On VBench, \method{} matches short-horizon baselines, achieves the best long-horizon overall score among evaluated methods, and improves throughput by $1.23\times$ on a single B200.
\end{itemize}

\section{Related Work}

\textbf{Causal Video Generation.}
Causal video diffusion converts a bidirectional teacher into a streaming student that generates frames or chunks autoregressively with a rolling KV cache. CausVid~\cite{yin2025slow} initiated this line with Distribution Matching Distillation (DMD) \cite{yin2024improved} based causal distillation, and Self-Forcing~\cite{huang2025self} reduced train--test mismatch by training on self-generated rollouts. Subsequent work improves long-horizon stability through joint denoising and attention sinks~\cite{liu2025rolling}, teacher-guided correction~\cite{cui2025self}, causal ODE initialization~\cite{zhu2026causal}, reward-weighted distillation and EMA sinks~\cite{lu2025reward}, deep sink and cache pruning~\cite{yi2025deep}, KV recaching for prompt switches~\cite{yang2025longlive}, and block-relative temporal RoPE~\cite{yesiltepe2025infinity}. These methods improve what is stored in the window or how it is positioned, but retain the dense per-head KV layout.

\textbf{Efficient Causal Video Generation.}
A complementary line restructures attention to reduce memory or compute. SANA-Video~\cite{chen2025sana} replaces softmax with block-causal linear attention and uses a constant-size cumulative state. SCD~\cite{bai2026causality} separates temporal reasoning from frame-wise rendering, caching only the causal encoder. VideoSSM~\cite{yu2025videossm} augments sliding-window KV with an SSM-compressed global memory. These approaches reduce temporal or layer-wise memory, but do not compress the per-token, per-head KV state at every cached layer.

\textbf{Multi-Head Latent Attention.}
DeepSeek-V2~\cite{liu2024deepseek} introduced Multi-Head Latent Attention (MLA), replacing per-head KV with a shared low-rank latent and a decoupled positional key; DeepSeek-V3~\cite{liu2024deepseekv3} scaled this design. MTLA~\cite{deng2025multi} further compresses along time, while MHA2MLA~\cite{ji2025towards} and TransMLA~\cite{meng2025transmla} convert pretrained MHA~\cite{vaswani2017attention} or GQA~\cite{ainslie2023gqa} LLMs into MLA. These works target language deployment. We study MLA in video diffusion, where the memory profile and pretrained attention spectrum differ substantially.

\section{Method}
\label{sec:method}
We write $x_t \in \mathbb{R}^d$ for the attention input at current chunk $t$, where a chunk denotes a group of latent frames. Let $d$ be the model dimension, $n_h$ the number of heads, and $d_h$ the per-head dimension, so that $d=n_hd_h$. VideoMLA introduces a shared KV latent dimension $d_c$ for cached content and splits each head into a NoPE content-scoring subspace and a RoPE positional subspace, $d_h=d_h^{\mathrm{nope}}+d_h^{\mathrm{rope}}$. The NoPE part is reconstructed from the shared latent and is not rotary-position encoded; the RoPE part uses a head-shared decoupled 3D-RoPE key.

\begin{figure}[t]
    \centering
        \includegraphics[width=1.0\linewidth]{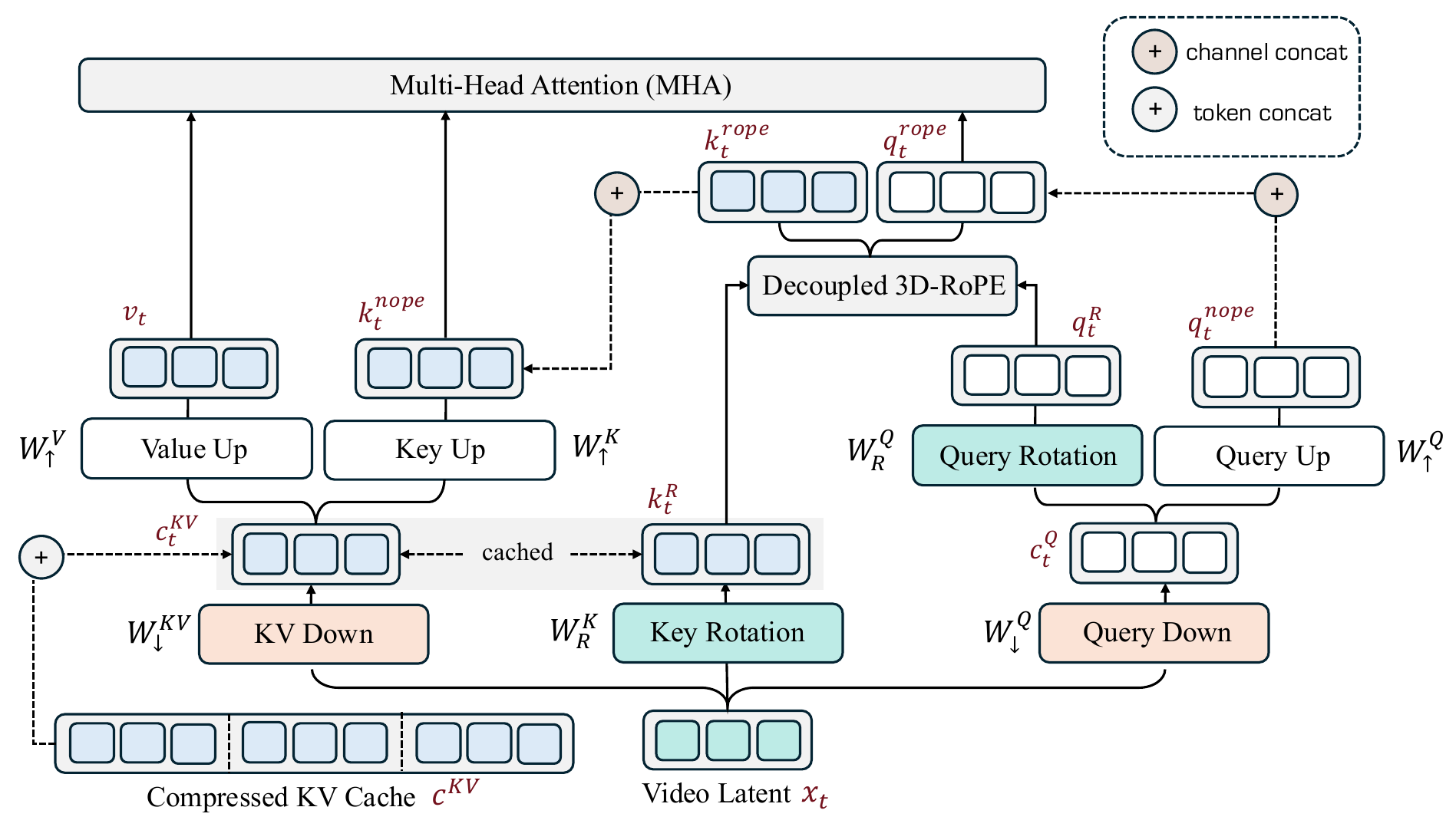} \\
    \caption{\textbf{Overview of \method{}.} \method{} replaces dense per-head KV cache in Causal Wan 2.1-1.3B with a low-rank latent obtained by jointly compressing keys and values through shared down/up projections, with positional information carried by a single decoupled rotated key. \textcolor{figOrange}{\textbf{Orange}} blocks denote down projections, \textcolor{figGreen}{\textbf{green}} blocks denote rotations, and white blocks denote up projections; latent frames are colored \textcolor{figBlue}{\textbf{blue}} for the key/value stream and white for the query stream. Each block is annotated with the corresponding weight matrix from Section~\ref{sec:method}, named latents are shown in \textcolor{red!70!black}{\textbf{red}}.}
\label{fig:framework}
\end{figure}

\subsection{Compressed KV Cache Construction}
\label{sec:method:cache}

Each video latent token $x_t \in \mathbb{R}^d$ produced by the backbone is first compressed into a low-rank latent that summarizes its key and value content for the rolling cache:
\begin{equation}
c_t^{KV} \;=\; W_\downarrow^{KV}x_t \;\in\; \mathbb{R}^{d_c},
\label{eq:ckv}
\end{equation}
where $W_\downarrow^{KV} \in \mathbb{R}^{d_c \times d}$ is the joint KV down-projection (Fig.~\ref{fig:framework}, KV Down). The vector $c_t^{KV}$ is the content object written into the compressed KV cache. It has dimension $d_c \ll n_hd_h$, so it replaces the dense per-head content keys and values that would otherwise be stored for every head. Positional information is not folded into this latent; it is stored separately through the decoupled key $k_t^R$ introduced in Section~\ref{sec:method:rope}. Thus, each cached token stores the pair $(c_t^{KV}, k_t^R)$ rather than dense per-head KV states.

The per-head keys and values needed by attention are obtained from $c_t^{KV}$ through two up-projections,
\begin{equation}
k_{t,h}^{\mathrm{nope}} \;=\; W_{\uparrow,h}^K\, c_t^{KV}, \qquad
v_{t,h} \;=\; W_{\uparrow,h}^V\, c_t^{KV},
\label{eq:reconstruct}
\end{equation}
where $h \in \{1, \dots, n_h\}$ indexes attention heads and $W_\uparrow^K, W_\uparrow^V$ are the key and value up-projections (Fig.~\ref{fig:framework}, Key Up and Value Up). Two properties of this construction are important. First, the same cached latent $c_t^{KV}$ is shared across all heads: a single cache read produces $n_h$ per-head keys and $n_h$ per-head values through Eq.~\ref{eq:reconstruct}. Second, the reconstructed key carries no rotary positional information. It is the content-only component of the per-head key, denoted $k_{t,h}^{\mathrm{nope}}$; the positional component lives in the separate RoPE subspace.

Together with the decoupled positional key, the per-token cached state is reduced from the $2n_hd_h$ scalars of a dense per-head KV cache to $d_c+d_h^{\mathrm{rope}}$ scalars. In our default setting, this is $224$ scalars per token per layer, a $92.7\%$ reduction.

The query path is per-token and uses an analogous down/up structure. From $x_t$, a query down-projection produces a query latent, and a content up-projection recovers the per-head NoPE query:
\begin{equation}
c_t^Q \;=\; W_\downarrow^Q\, x_t \;\in\; \mathbb{R}^{d_q},
\label{eq:cq}
\end{equation}
\begin{equation}
q_{t,h}^{\mathrm{nope}} \;=\; W_{\uparrow,h}^Q\, c_t^Q,
\label{eq:qnope}
\end{equation}
where $d_q$ is the query latent dimension and $W_\uparrow^Q$ is the Query Up projection in Fig.~\ref{fig:framework}. Since queries are recomputed from the current block at every generation step, $c_t^Q$ is internal to the layer and is never written to the KV cache. The head-sharing occurs only in the decoupled positional branch: \method{} uses a single RoPE key shared across heads, while the NoPE queries, NoPE keys, and values remain head-specific after up-projection.

The dimension $d_c$ is the layer's main content-cache capacity knob: it controls how aggressively the cached content is compressed and how much shared subspace the model can use for joint key-value content. The choice of $d_c$ is studied empirically in Figure \ref{fig:oom_sweep} and Appendix.

\begin{table}[t]
\centering
\vspace{-0.5em}
\label{tab:attention-cost}
\small
\setlength{\tabcolsep}{6pt}
\renewcommand{\arraystretch}{1.35}
\begin{tabularx}{\linewidth}{X|ccc|c}
\toprule
\rowcolor{verylightgray} %
\textbf{Metric} &
\textbf{Causal Full} &
\textbf{Causal Local} &
\textbf{Causal Linear} &
\textbf{MLA Local} \\
\midrule
Memory
  & $2ND$
  & $2WD$
  & $D\, d_h$
  & $W\!\left(d_c + d_h^{\mathrm{rope}}\right)$ \\
Comp.\ ($N$-th token)
  & $ND$
  & $WD$
  & $D\, d_h$
  & $n W\!\left(d_c + d_h^{\mathrm{rope}}\right)$ \\
Comp.\ ($N$ tokens)
  & $\tfrac{1}{2} N^{2} D$
  & $N W D$
  & $N\, D\, d_h$
  & $n N W\!\left(d_c + d_h^{\mathrm{rope}}\right)$ \\
\bottomrule
\end{tabularx}
\vspace{0.5em}
\caption{\textbf{Memory and compute costs across four attention variants.} For a sequence of length $N$ with hidden dimension $D$, $n$ heads, per-head dimension $d_h = D/n$, local window $W$, latent KV dimension $d_c$ ($d_c \ll D$), and shared decoupled-RoPE dimension $d_h^{\mathrm{rope}}$.}
\end{table}

\subsection{Decoupled 3D-RoPE}
\label{sec:method:rope}

The latent cache $c_t^{KV}$ is kept position-free, so that the low-rank content path can be shared across heads and reused under sliding-window re-indexing. Positional information is instead carried by a separate RoPE subspace. We split each head as
$d_h=d_h^{\mathrm{nope}}+d_h^{\mathrm{rope}}$, where $k_{t,h}^{\mathrm{nope}}$ is the reconstructed content key and the remaining channels form a decoupled 3D-RoPE key. As in Wan, $d_h^{\mathrm{rope}}$ is partitioned across temporal, height, and width axes, using the corresponding high-frequency rotary bands.

For each token, \method{} computes a single head-shared positional key
\begin{equation}
k_t^R = W_R^K x_t \in \mathbb{R}^{d_h^{\mathrm{rope}}},
\qquad
k_t^{\mathrm{rope}} = \mathrm{RoPE}_{3D}(k_t^R),
\label{eq:krope}
\end{equation}
rather than $n_h$ per-head RoPE keys. The cache stores the unrotated state $(c_t^{KV}, k_t^R)$; rotation is applied only when the active attention window is assembled. This keeps cached states independent of absolute rollout time and yields a per-token cache size of $d_c+d_h^{\mathrm{rope}}$.

The query branch follows the same decomposition. From the query latent $c_t^Q$, the positional query for head $h$ is
\begin{equation}
q_{t,h}^R = W_{R,h}^Q c_t^Q,
\qquad
q_{t,h}^{\mathrm{rope}} = \mathrm{RoPE}_{3D}(q_{t,h}^R).
\label{eq:qrope}
\end{equation}
Attention is then computed over the concatenated NoPE and RoPE components: each head uses $(q_{t,h}^{\mathrm{nope}}, q_{t,h}^{\mathrm{rope}})$ against $(k_{t,h}^{\mathrm{nope}}, k_t^{\mathrm{rope}})$, while values remain reconstructed only from the content latent.

\subsection{Training-Time Forward Pass}
\label{sec:method:training}

During training, every video latent token writes its compressed cache state $(c_t^{KV}, k_t^R)$, defined in Eqs.~\ref{eq:ckv} and \ref{eq:krope}, into the KV cache as the block under denoising progresses. Attention is then computed in standard multi-head form, with the per-head content keys and values reconstructed on demand from the cached content latent through Eq.~\ref{eq:reconstruct}, and the shared positional key obtained by rotating the cached positional state $k_t^R$ at use time.

For a query token at position $i$ and a cached token at position $j$, attention head $h$ combines the content and positional contributions into a single score
\begin{equation}
\mathrm{score}^{(h)}_{i, j} \;=\; \frac{q_{i,h}^{\mathrm{nope}} \cdot k_{j,h}^{\mathrm{nope}} \;+\; q_{i,h}^{\mathrm{rope}} \cdot k_j^{\mathrm{rope}}}{\sqrt{d_h^{\mathrm{nope}} + d_h^{\mathrm{rope}}}},
\label{eq:score}
\end{equation}
where $q_{i,h}^{\mathrm{nope}}$ and $q_{i,h}^{\mathrm{rope}}$ are the content and rotated positional query components from Eqs.~\ref{eq:qnope} and \ref{eq:qrope}, $k_{j,h}^{\mathrm{nope}}$ is the per-head content key from Eq.~\ref{eq:reconstruct}, and $k_j^{\mathrm{rope}}$ is the rotated shared positional key obtained from Eq.~\ref{eq:krope}. The two inner products live in subspaces of dimension $d_h^{\mathrm{nope}}$ and $d_h^{\mathrm{rope}}$ respectively, so the joint score is normalized by their combined dimension. A softmax over the active attention window followed by a weighted sum of $v_{j,h}$ produces the per-head output, and the head outputs are mixed through the output projection $W^O$.

The shape of $\mathrm{score}^{(h)}_{i,j}$ matches what a dense attention layer of the same per-head dimension would produce. As a consequence, \method{} substitutes for the dense self-attention module without any change to the surrounding training pipeline: chunkwise causal block masks, sink tokens, and FlexAttention kernels operate on the reconstructed per-head keys and values exactly as they would on dense ones. The only structural change relative to the dense baseline is internal to the attention layer: the cache holds $(c_t^{KV}, k_t^R)$ rather than per-head $K$ and $V$, and the per-head views consumed by attention are reconstructed at use time.

\section{Experiments}
\label{sec:experiments}
\subsection{Setup and Dataset}

\textbf{Implementation Details.} We implement \method{} on top of the Wan-2.1 T2V-1.3B backbone~\cite{wan2025wan}, replacing only the self-attention layers while leaving the remaining architecture unchanged. The model has $30$ transformer blocks, hidden dimension $1536$, $12$ heads, and per-head dimension $128$. Unless otherwise stated, we use $d_c=192$ and $d_q=768$, with the head dimension split into $d_h^{\mathrm{nope}}=96$ and $d_h^{\mathrm{rope}}=32$. The decoupled 3D-RoPE channels are allocated across temporal, height, and width axes as $(6,5,5)$ complex pairs, using the highest-frequency bands. This gives a per-token cache size of $d_c+d_h^{\mathrm{rope}}=224$ scalars, corresponding to a $13.7\times$ reduction from the dense $2n_hd_h=3072$-scalar KV cache. Training follows the three-stage Causal Forcing pipeline~\cite{zhu2026causal}, including Teacher Forcing, Consistency Distillation initialization to four steps, and DMD, with total batch size $128$. We use learning rates $5{\times}10^{-6}$ for Teacher Forcing and $2{\times}10^{-6}$ for Consistency Distillation and DMD. All training experiments are run in bf16 mixed precision on a 8 $\times$ B200 GPU.

\textbf{Dataset.}
For the Consistency Distillation stage preceding DMD, we use 47{,}680 videos: 29{,}471 from OpenVid-1M~\cite{nan2024openvid} and 18{,}209 synthesized clips. 
\begin{figure}[t]
    \centering
        \includegraphics[width=1.0\linewidth]{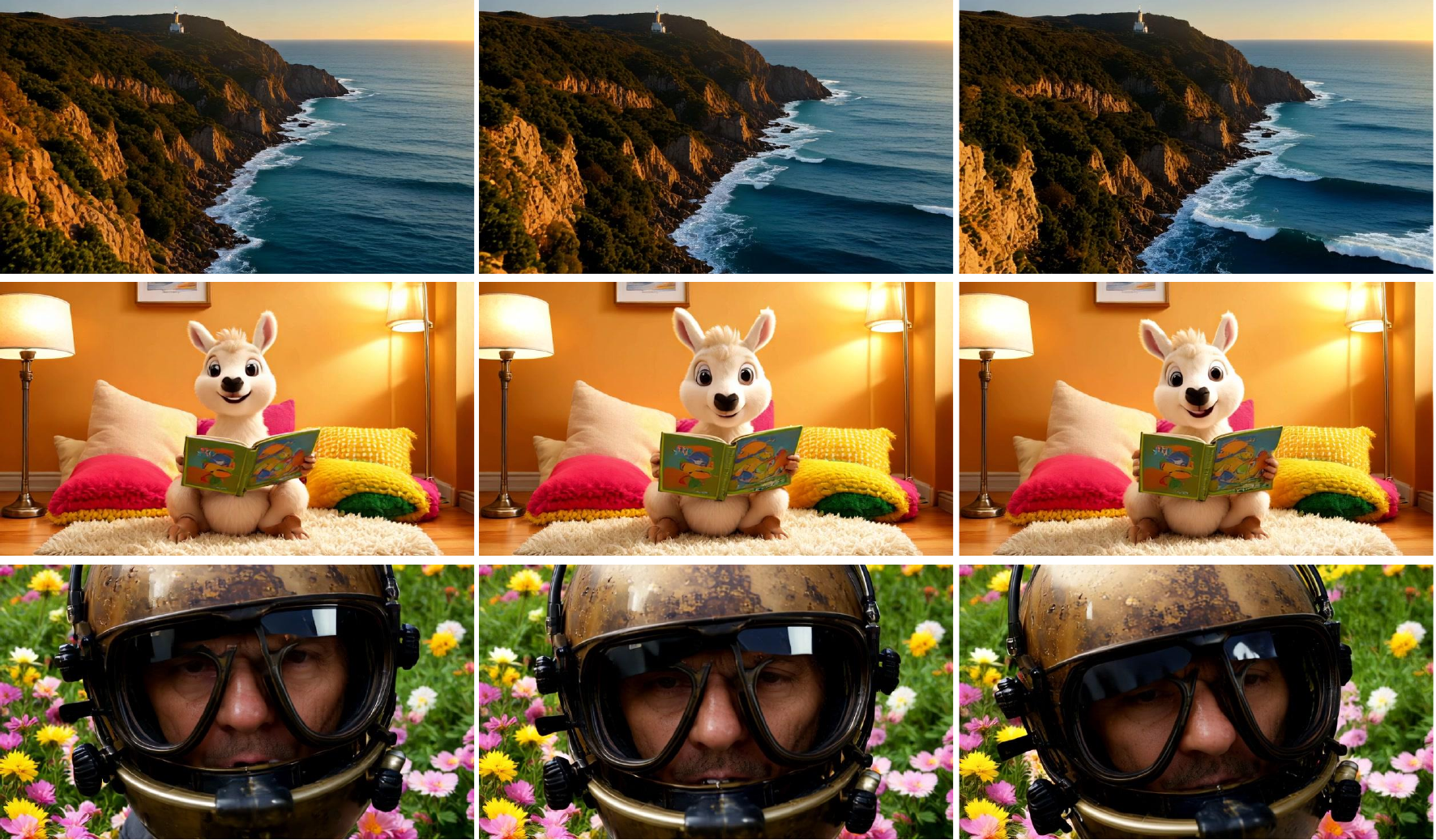} \\
    \caption{\textbf{Qualitative results.}
Samples generated by \method{}. Frames are shown at uniformly spaced timestamps from each 30s rollout, illustrating that the compressed latent KV cache preserves scene structure, subject identity, and visual fidelity over time.}
\label{fig:qualitative}
\end{figure}
\begin{figure}[t]
    \centering
        \includegraphics[width=1.0\linewidth]{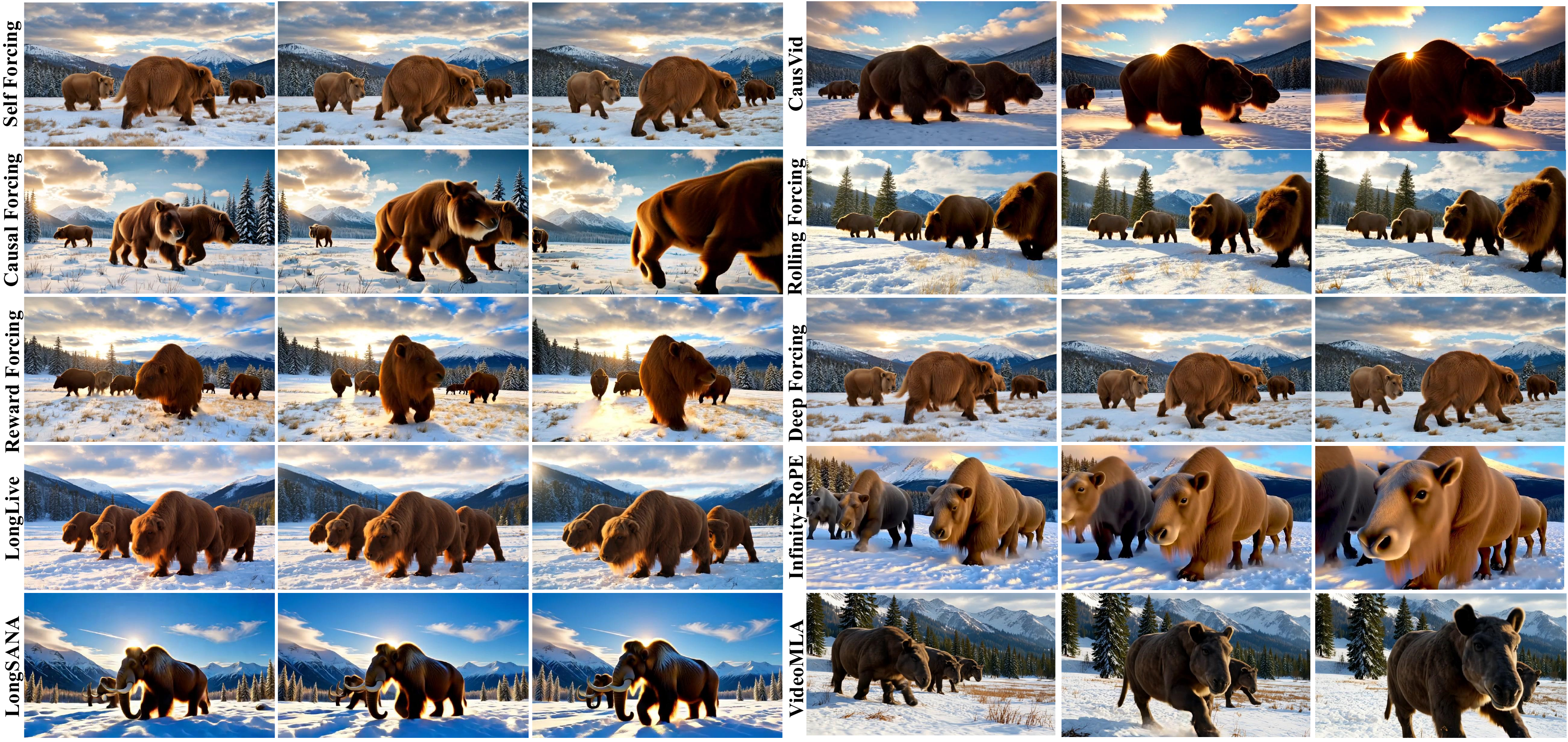} \\
    \caption{\textbf{Qualitative comparison.}
Long-rollout samples from \method{} and baseline causal video diffusion baselines under the same prompt. Each row shows uniformly spaced frames from one method.}
\label{fig:qual_comparison}
\end{figure}

\textbf{Baselines.}
We compare \method{} with recent causal video diffusion methods covering standard streaming pipelines, attention-architecture redesigns, and positional reparameterizations. The streaming baselines include CausVid~\citep{yin2025slow}, Self-Forcing~\citep{huang2025self}, Rolling-Forcing~\citep{liu2025rolling}, Causal Forcing~\citep{zhu2026causal}, Reward Forcing~\citep{lu2025reward}, Deep Forcing~\citep{yi2025deep}, and LongLive~\citep{yang2025longlive} and Infinity-RoPE\citep{yesiltepe2025infinity}. We also compare with architectural efficiency method LongSANA~\citep{chen2025sana}.

\subsection{Main Results}
\textbf{Qualitative Results.} Figure~\ref{fig:qualitative} shows that \method{} preserves subject identity, scene structure, and visual fidelity over 30-second rollouts despite replacing the dense per-head KV cache with a compact latent cache. Finally, Figure~\ref{fig:qual_comparison} shows that \method{} generates results comparable to representative streaming causal video baselines while requiring faster inference and substantially lower memory. These qualitative results indicate that \method{} improves the efficiency--memory trade-off without the pronounced fidelity, dynamism, or long-horizon stability losses observed in more aggressive compression-based alternatives.

\textbf{Quantitative Results.} Table~\ref{tab:res_long_horizon} reports long-horizon VBench results at 30s and 60s. \method{} achieves the best dynamic degree at both horizons, with 0.981 at 30s and 0.958 at 60s, indicating that latent KV compression does not suppress motion or lead to static generation. It also obtains the best imaging quality and motion smoothness, and reaches the highest 60s overall score of 0.859, substantially outperforming prior streaming baselines such as Reward Forcing, Infinity-RoPE, LongLive, and LongSANA. At 30s, \method{} is also competitive with the strongest baseline, achieving the second-best overall score while using a much smaller KV cache memory.

\begin{table*}[t]
\centering
\scriptsize
\setlength{\tabcolsep}{3pt}
\renewcommand{\arraystretch}{1.15}
\resizebox{\textwidth}{!}{
\begin{tabular}{l cccccc c !{\vrule} cccccc c !{\vrule} cccc}
\toprule
\multirow{2}{*}{\textbf{Model}} &
\multicolumn{7}{c}{\cellcolor{verylightgray}\textbf{Results on 30s}\,$\uparrow$} &
\multicolumn{7}{c}{\cellcolor{verylightgray}\textbf{Results on 60s}\,$\uparrow$} &
\multicolumn{4}{c}{\cellcolor{verylightgray}\textbf{User Study}\,$\uparrow$} \\
\cmidrule(lr){2-8} \cmidrule(lr){9-15} \cmidrule(lr){16-19}
&
\textbf{AQ} & \textbf{BC} & \textbf{DD} & \textbf{IQ} & \textbf{MS} & \textbf{SC} & \textbf{Overall} &
\textbf{AQ} & \textbf{BC} & \textbf{DD} & \textbf{IQ} & \textbf{MS} & \textbf{SC} & \textbf{Overall} &
\textbf{PA} & \textbf{TC} & \textbf{DC} & \textbf{Overall} \\
\midrule
Self-Forcing~\cite{huang2025self}
& 0.541 & 0.948 & 0.624 & 0.577 & 0.952 & 0.932 & 0.762
& 0.565 & 0.958 & 0.393 & 0.650 & 0.987 & 0.974 & 0.755
& 2.79 & 2.79 & 2.70 & 2.76 \\
CausVid~\cite{yin2025slow}
& 0.597 & 0.921 & 0.473 & 0.663 & 0.935 & 0.913 & 0.750
& 0.497 & 0.929 & 0.723 & 0.574 & 0.948 & 0.933 & 0.767
& -- & -- & -- & -- \\
Causal Forcing~\cite{zhu2026causal}
& 0.526 & 0.945 & 0.738 & 0.628 & 0.968 & 0.947 & 0.792
& 0.503 & 0.936 & \underline{0.847} & 0.608 & 0.935 & 0.920 & 0.792
& 2.59 & 2.63 & \underline{2.81} & 2.68 \\
Rolling-Forcing~\cite{liu2025rolling}
& 0.620 & 0.953 & 0.742 & \underline{0.688} & 0.982 & 0.960 & 0.824
& 0.580 & 0.958 & 0.380 & 0.670 & 0.988 & 0.977 & 0.759
& 2.55 & 2.68 & 2.60 & 2.61 \\
Deep Forcing~\cite{yi2025deep}
& 0.621 & 0.953 & 0.713 & 0.660 & 0.979 & 0.961 & 0.815
& 0.597 & 0.957 & 0.402 & 0.690 & 0.987 & 0.979 & 0.769
& 2.60 & 2.76 & 2.68 & 2.68 \\
Reward Forcing~\cite{lu2025reward}
& \underline{0.644} & 0.956 & \underline{0.954} & 0.683 & 0.981 & 0.957 & \textbf{0.863}
& 0.585 & 0.952 & 0.676 & 0.673 & 0.985 & 0.974 & \underline{0.808}
& \underline{2.91} & \underline{2.99} & 2.83 & \underline{2.91} \\
LongLive~\cite{yang2025longlive}
& \textbf{0.654} & \underline{0.959} & 0.649 & 0.678 & \underline{0.983} & \underline{0.967} & 0.816
& \underline{0.606} & 0.961 & 0.433 & 0.664 & \underline{0.991} & \underline{0.982} & 0.773
& 2.56 & 2.70 & 2.58 & 2.61 \\
Infinity-RoPE~\cite{yesiltepe2025infinity}
& 0.640 & 0.958 & 0.847 & 0.669 & 0.982 & 0.966 & 0.844
& \textbf{0.607} & 0.959 & 0.647 & 0.638 & 0.988 & 0.979 & 0.803
& 2.46 & 2.44 & 2.41 & 2.43 \\
\midrule
LongSANA~\cite{chen2025sana}
& 0.573 & \textbf{0.976} & 0.149 & 0.683 & 0.974 & \textbf{0.988} & 0.723
& 0.529 & \textbf{0.976} & 0.103 & \underline{0.702} & \underline{0.991} & \textbf{0.986} & 0.714
& 2.48 & 2.63 & 2.56 & 2.56 \\
\rowcolor{lightblue}
\textbf{\method{} (Ours)}
& 0.601 & 0.942 & \textbf{0.981} & \textbf{0.697} & \textbf{0.986} & 0.952 & \underline{0.859}
& 0.569 & \underline{0.963} & \textbf{0.958} & \textbf{0.715} & \textbf{0.993} & 0.954 & \textbf{0.859}
& \textbf{3.04} & \textbf{3.24} & \textbf{3.22} & \textbf{3.17} \\
\bottomrule
\end{tabular}
}
\caption{\textbf{Long-horizon performance and user preference comparison.} Results across 30s and 60s video generation, plus user study scores. \textbf{AQ}: Aesthetic Quality, \textbf{BC}: Background Consistency, \textbf{DD}: Dynamic Degree, \textbf{IQ}: Imaging Quality, \textbf{MS}: Motion Smoothness, \textbf{SC}: Subject Consistency. User study metrics are \textbf{PA}: Prompt Adherence, \textbf{TC}: Temporal Consistency, and \textbf{DC}: Dynamic Consistency. \textbf{Bold}: best; \underline{underline}: second best.}
\label{tab:res_long_horizon}
\vspace{-0.5em}
\end{table*}
\begin{figure}[t]
    \centering
        \includegraphics[width=1.0\linewidth]{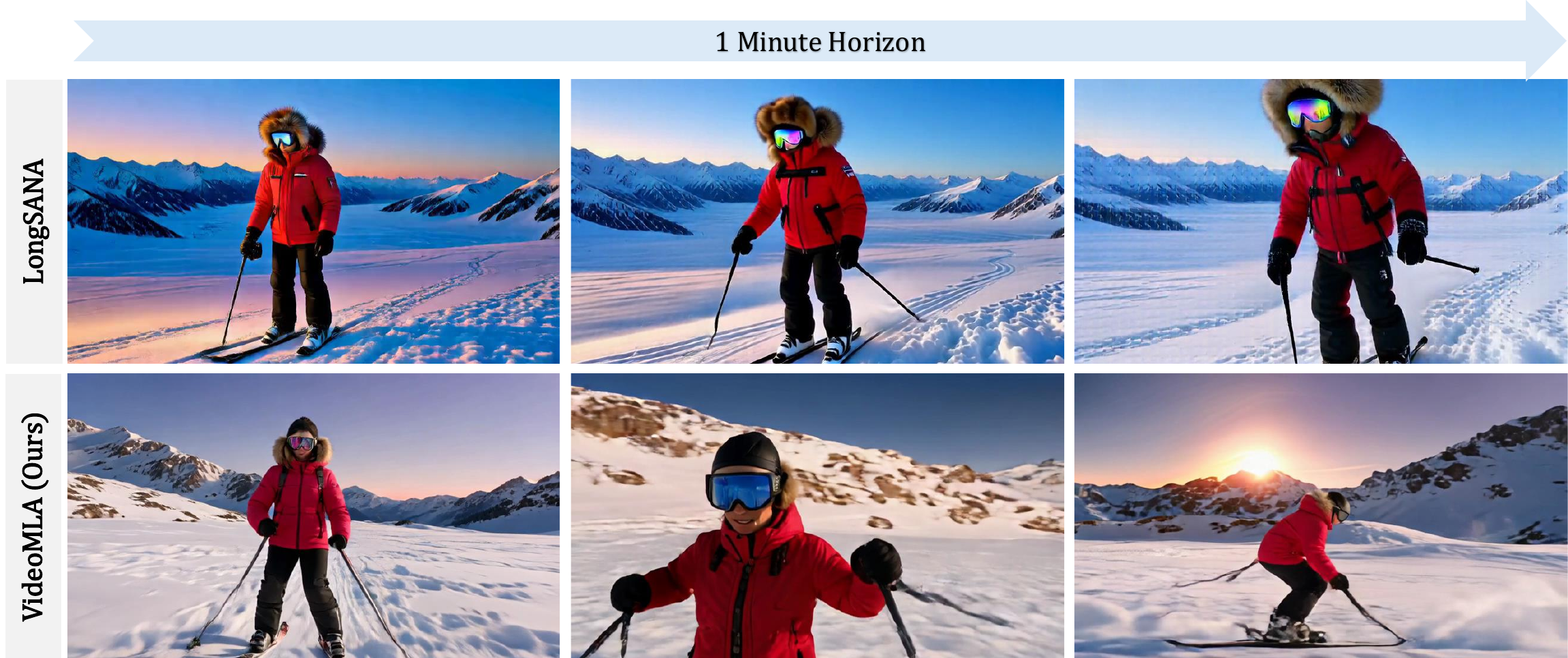} \\
    \caption{\textbf{Long-horizon generation quality.} Frames sampled across a one-minute rollout of the same prompt. \textbf{(Bottom)} \method{}  sustains visual fidelity with diverse, evolving motion, while \textbf{(Top)} LongSANA produces near-static content that degrades over time. \method{} yields higher visual fidelity and more diverse motion while achieving higher generation throughput and lower latency than LongSANA, and reduces KV cache size by $92.7\%$ relative to the Self-Forcing baseline.}
\label{fig:LongSana_comparison}
\end{figure}

\begin{table}[t]
\centering
\small
\setlength{\tabcolsep}{5pt}
\renewcommand{\arraystretch}{1.1}
\begin{tabularx}{\linewidth}{@{}>{\raggedright\arraybackslash}X cccc|ccc@{}}
\toprule
\textbf{Model} &
\textbf{\#Params} &
\textbf{Resolution} &
\textbf{Throughput}\,$\uparrow$ &
\textbf{Latency}\,$\downarrow$ &
\textbf{CLIP-T}\,$\uparrow$ &
\textbf{CLIP-F}\,$\uparrow$ &
\textbf{HPSv3}\,$\uparrow$ \\
\midrule
\rowcolor{lightblue}
\multicolumn{8}{l}{\textit{Frame-wise autoregressive models}} \\
NOVA~\cite{deng2024autoregressive}             & 0.6B & 768$\times$480 & 2.26   & 14.63   & 0.2764 & 0.9673 & 2.95 \\
\mbox{Pyramid Flow}~\cite{jin2024pyramidal}    & 2B   & 640$\times$384 & 1.39   & 87.32  & 0.2888 & 0.9795 & 8.02 \\
\midrule
\rowcolor{lightblue}
\multicolumn{8}{l}{\textit{Chunk-wise autoregressive models}} \\
Self-Forcing~\cite{huang2025self}              & 1.3B & 832$\times$480 & 18.06           & 4.19          & 0.3036 & \underline{0.9689} & \textbf{9.86} \\
LongSANA~\cite{chen2025sana}                   & 2B   & 832$\times$480 & \underline{19.35} & \underline{4.48} & \underline{0.2978} & \textbf{0.9887} & 7.54 \\
\textbf{\method{}}~(Ours)                      & 1.3B & 832$\times$480 & \textbf{23.96} & \textbf{3.38} & \textbf{0.3278}          & 0.9686          & \underline{9.74} \\
\bottomrule
\end{tabularx}
\vspace{0.5em}
\caption{\textbf{Text-to-Video quantitative comparison on VBench.} Models have similar parameter sizes and resolutions. Throughput\,$\uparrow$ (FPS) and latency\,$\downarrow$ (s) measured with batch size 1 on \textbf{B200}. Higher is better for CLIP-T, CLIP-F, and HPSv3 scores\,$\uparrow$. \textbf{Bold}: best; \underline{underline}: second best.}
\label{tab:efficiency-t2v}
\end{table}
\textbf{Efficiency Results.} Table~\ref{tab:efficiency-t2v} shows that \method{} achieves the highest throughput and lowest latency among chunk-wise autoregressive models, while also obtaining the best CLIP-T score. Although LongSANA has a slightly higher CLIP-F score, this is partly due to its more static generations, which preserve frame-level similarity but reduce motion dynamics. Consistently, \method{} obtains a higher HPSv3 score and, as shown in Figure~\ref{fig:LongSana_comparison}, produces sharper, more dynamic, and more temporally stable long-rollout videos than LongSANA.

\subsection{Ablations}
\label{sec:ablations}

\begin{figure}
    \centering
        \includegraphics[width=1.0\linewidth]{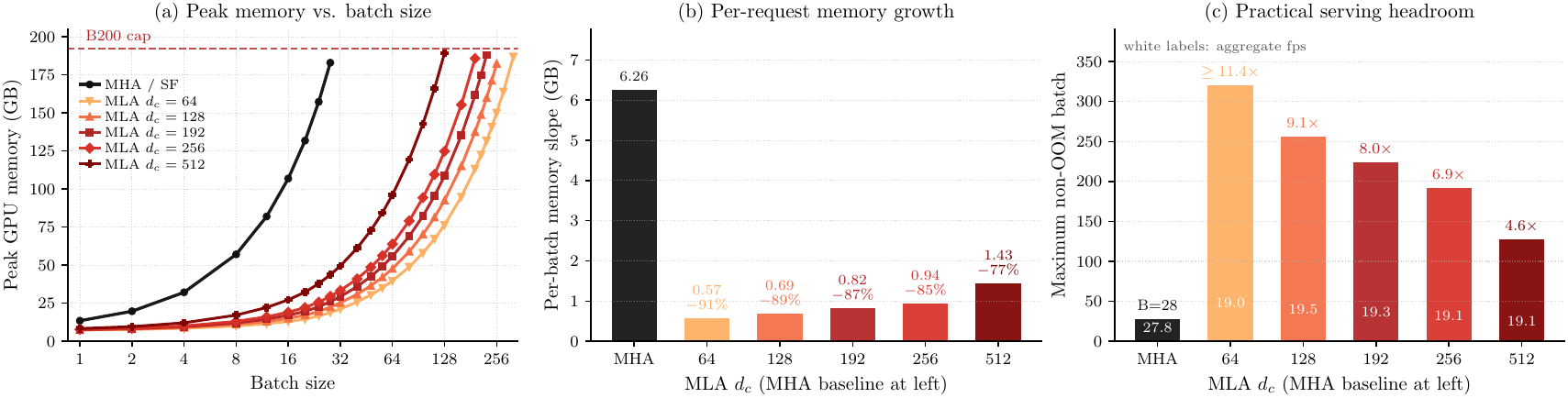} \\
    \caption{\textbf{\method{} increases serving headroom under a fixed B200 memory budget.}
Compared with dense MHA, MLA greatly reduces per-batch memory growth and shifts the OOM limit to much larger batch sizes; the default $d_c=192$ gives $8.0\times$ non-OOM batch headroom.}
\label{fig:oom_sweep}
\end{figure}

\textbf{Batch Scaling Under Fixed Memory.} Fig.~\ref{fig:oom_sweep} shows that \method{} translates cache compression into practical serving headroom on a single B200. Dense MHA reaches the memory limit at $B=28$, whereas MLA shifts the OOM cliff far to the right; with $d_c=64$, it remains within budget even at $B=320$. The per-request memory slope drops from $6.26$ GB/batch for MHA to $0.57$--$1.43$ GB/batch for MLA, a $77$--$91\%$ reduction across $d_c \in \{64,128,192,256,512\}$. Consequently, MLA supports $4.6\times$ to at least $11.4\times$ larger non-OOM batches under the same memory cap, with our default $d_c=192$ giving $8.0\times$ batch headroom.

\section{Why MLA Works in Video Diffusion: Rank Budget vs. Spectral Structure}
\label{sec:spectral_analysis}

MLA is often motivated by the assumption that the pretrained key/value maps are approximately low-rank. We test whether this explanation holds for video diffusion by analyzing the joint dense operator $[W_K;\,W_V]$ in Wan2.1-T2V-1.3B. Fig.~\ref{fig:svd_spectrum} shows that this operator is not low-rank: at the default budget $d_c=192$, the median layer preserves only $45.8\%$ of the spectral energy, and the 99\%-energy effective rank exceeds 1300 in every layer. Thus, a direct rank-$d_c$ spectral approximation would discard most of the dense key/value energy, even though \method{} retains generation quality at this cache size.

This mismatch suggests that MLA should not be interpreted as recovering a hidden low-rank structure in the pretrained attention weights. Instead, MLA changes the optimization problem: the composed key/value operator
\[
M = [W^K_\uparrow W^{KV}_\downarrow;\, W^V_\uparrow W^{KV}_\downarrow]
\]
is constrained by construction to have rank at most $d_c$. Fig.~\ref{fig:learned_rank} confirms that the learned composed operator uses this architectural budget almost fully across latent sizes. For $d_c \in \{64,128,256,512\}$, the normalized spectra share a common shape truncated at $d_c$, and the layer-wise 99\%-energy rank remains close to $0.98d_c$ throughout the network. The effective rank is therefore set by the MLA bottleneck rather than by the spectrum of the original dense operator.

\begin{figure}
    \centering
        \includegraphics[width=1.0\linewidth]{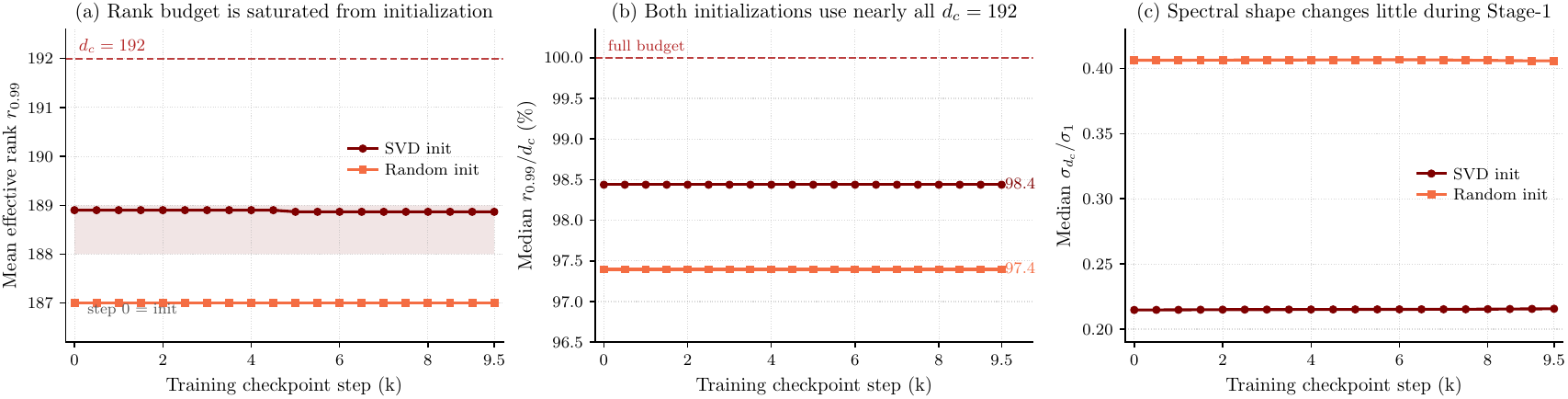} \\
    \caption{\textbf{Rank-budget saturation during training.}
    At $d_c=192$, both SVD and random initialization occupy nearly the full latent rank budget from initialization, with stable effective rank and spectral tail throughout training.}
    \label{fig:rank_budget_stage1}
\end{figure}
We further investigate whether this is an artifact of SVD initialization. Fig.~\ref{fig:rank_budget_stage1} compares SVD and random initialization at $d_c=192$ during training. Both nearly saturate the rank budget from initialization, and training preserves the effective rank and spectral tail. Thus, training does not discover a lower-rank solution or collapse the spectrum; it adapts within the imposed budget.

\section{Limitations and Broader Impact}

VideoMLA reduces per-token KV cache, but the latent budget cannot shrink arbitrarily. Small budgets such as $d_c=64$ improve memory headroom but lose fine-grained details and degrade quality, making $d_c$ a quality--efficiency trade-off. Our experiments focus on Wan2.1-T2V-1.3B and minute-scale generation; larger backbones, higher resolutions, longer horizons, and prompt switching remain future work. More efficient long-horizon generation can reduce deployment cost and broaden access to creative tools, simulation, education, and assistive media production.

\section{Conclusion}

We presented VideoMLA, the first MLA-style latent KV cache for autoregressive video diffusion. By replacing dense per-head keys and values with a shared low-rank content latent and a head-shared decoupled 3D-RoPE positional key, VideoMLA reduces per-token KV cache memory by 92.7\% while preserving compatibility with standard chunk-causal generation. Our analysis shows that this success does not arise from an intrinsically low-rank pretrained key-value operator; instead, the MLA bottleneck defines a rank budget that the model uses nearly fully and adapts within during training. Empirically, VideoMLA preserves visual quality and motion at long horizons, achieves the best one-minute overall score among evaluated methods, and improves throughput with substantially lower cache memory. These results identify the per-token KV layout as an effective and complementary axis for scaling efficient long-horizon video diffusion.

\section*{Acknowledgements}

Pinar Yanardag is supported by the National Science Foundation
under Grant No.\ 2543524.

\bibliographystyle{splncs04}
\bibliography{main}

\clearpage
\setcounter{page}{1}
\appendix

\section*{Table of Contents}
\addcontentsline{toc}{section}{Supplementary Material Table of Contents}
\startcontents[appendix]
\printcontents[appendix]{l}{1}{\setcounter{tocdepth}{2}}

\section{Videos and Website}
\label{sec:website}
To facilitate comprehensive evaluation and improve result accessibility, we provide video results covering qualitative examples, ablation studies, comparisons, and limitations in the 
\normalsize{\url{https://videomla.github.io}}.

\section{Details on User Study}
\begin{figure}
    \centering
        \includegraphics[width=1.0\linewidth]{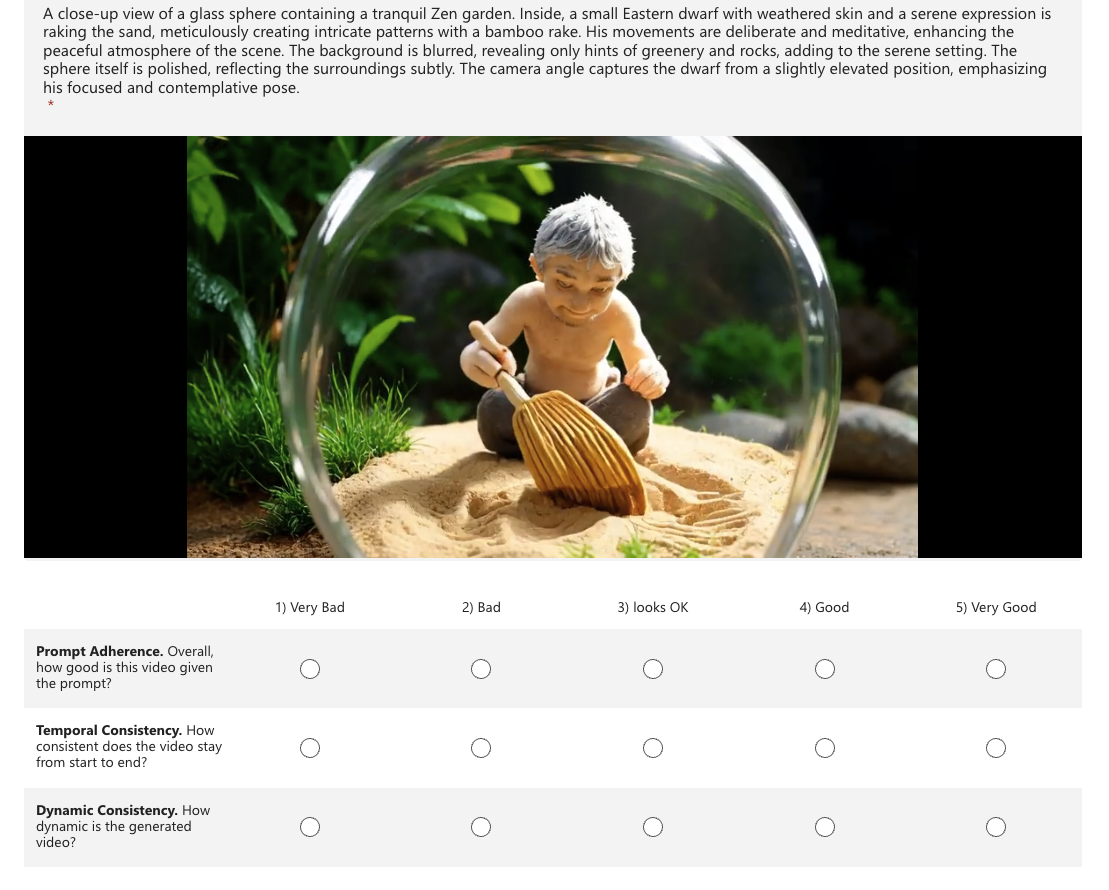} \\
    \caption{\textbf{User Study Interface.} User Study Interface for Long Video Generation}
\label{fig:user_study}
\end{figure}

We conduct a user study to evaluate perceptual quality of one-minute generations. We compare nine models and ask 50 participants to rate each video using the interface shown in Fig.~\ref{fig:user_study}. For each generated video, participants answer three questions: \emph{Prompt Adherence}, measuring how well the video follows the prompt; \emph{Temporal Consistency}, measuring whether the video remains coherent from start to end; and \emph{Dynamic Consistency}, measuring whether the video contains plausible and sustained motion. Each question is rated on a five-point Likert scale, from 1) Very Bad to 5) Very Good. 

\section{Background}
\subsection{Wan2.1-T2V-1.3B Backbone}

Our experiments use Wan2.1-T2V-1.3B as the base video diffusion backbone. Wan2.1-T2V-1.3B is a latent video diffusion transformer operating over spatiotemporal latent tokens rather than RGB pixels. The video is encoded by a 3D VAE that compresses the temporal dimension by $4\times$ and each spatial dimension by $8\times$, so an input video $V \in \mathbb{R}^{F \times H \times W \times 3}$ is mapped to a latent tensor with temporal length $1 + \lceil (F-1)/4 \rceil$ and spatial resolution $H/8 \times W/8$. The denoising model follows the rectified-flow formulation, where a clean latent $x_0$ and Gaussian noise $\epsilon$ are linearly interpolated as
\[
x_t = (1-t)x_0 + t\epsilon,\qquad t\in[0,1],
\]
and the reverse process is parameterized by a neural velocity field and solved with Euler integration at inference time.

Wan2.1-T2V-1.3B uses a diffusion transformer with multi-head self-attention over video latent tokens. In our implementation, the backbone contains 30 transformer blocks, hidden dimension $d=1536$, $n_h=12$ attention heads, and per-head dimension $d_h=128$. In the dense baseline, each cached token stores both keys and values for all heads, giving a per-token, per-layer KV cache size of
\[
2n_hd_h = 2 \times 12 \times 128 = 3072
\]
scalars. With a 21-latent-frame cache, 1,560 tokens per latent frame, and 30 cached transformer layers, this corresponds to 3.02B cached scalars, or approximately 6.0GB in bf16/fp16. This dense per-head KV layout is the main memory target of VideoMLA.

The backbone uses 3D rotary position embeddings (3D-RoPE) to encode temporal and spatial token coordinates before self-attention. For latent features $x \in \mathbb{R}^{B \times S \times C}$ with $S=FHW$, the channel dimension is partitioned across temporal, height, and width axes, and RoPE is applied separately to the corresponding coordinate subspaces before concatenation. In Wan, each RoPE dimension has a fixed maximum sequence length of 1024; although RoPE remains mathematically defined beyond this range, generation outside the positional regime observed during training can degrade attention quality.

For autoregressive long-video generation, Wan2.1-T2V-1.3B is commonly used after causal distillation or self-rollout training. The model generates latent frame chunks sequentially and conditions each chunk on a rolling KV cache of previous chunks. This cache enables efficient streaming generation because previous key and value states are reused rather than recomputed. However, in the dense Wan attention layout, the cache still stores full per-head keys and values for every retained token and every cached layer. VideoMLA keeps the Wan backbone and causal rollout setting intact, but replaces this dense per-token KV state with a shared latent content cache and a decoupled head-shared 3D-RoPE key.

\begin{table}[t]
\centering
\caption{\textbf{Training and model hyperparameters.} Unless otherwise stated, all experiments use the default \method{} setting.}
\label{tab:hyperparameters}
\small
\begin{tabular}{l c}
\toprule
\textbf{Hyperparameter} & \textbf{Value} \\
\midrule
Backbone & Wan2.1-T2V-1.3B \\
Transformer blocks & 30 \\
Hidden dimension $d$ & 1536 \\
Number of heads $n_h$ & 12 \\
Per-head dimension $d_h$ & 128 \\
KV latent dimension $d_c$ & 192 \\
Query latent dimension $d_q$ & 768 \\
NoPE channels $d_h^{\mathrm{nope}}$ & 96 \\
RoPE channels $d_h^{\mathrm{rope}}$ & 32 \\
3D-RoPE complex pairs $(t,h,w)$ & $(6,5,5)$ \\
Per-token cache size & 224 scalars \\
KV cache reduction & $92.7\%$ \\
Training precision & bf16 \\
Training GPUs & $8\times$ B200 \\
Total batch size & 128 \\
Teacher Forcing LR & $5\times 10^{-6}$ \\
Consistency Distillation LR & $2\times 10^{-6}$ \\
DMD LR & $2\times 10^{-6}$ \\
Training pipeline & TF $\rightarrow$ CD $\rightarrow$ DMD \\
Inference steps & 4 \\
\bottomrule
\end{tabular}
\end{table}

\section{Implementation Details}
\label{app:implementation_details}

\subsection{Backbone and Tokenization}
\label{app:backbone_tokenization}

VideoMLA is implemented on top of Wan2.1-T2V-1.3B. The backbone contains
$L=30$ transformer blocks, hidden dimension $d=1536$, $n_h=12$ attention heads,
and per-head dimension $d_h=128$, so that $d=n_h d_h$. The feed-forward hidden
dimension is $8960$. We keep the Wan text-conditioning branch and all non-attention
modules unchanged, and replace only the temporal self-attention layers with the
VideoMLA block described in Section~3.

We train on 5-second clips at $480\times832$ resolution and 16 fps. The Wan VAE
encodes each clip into a latent tensor with 16 channels, 21 latent frames, and
spatial size $60\times104$. A 3D patch embedding with patch size $(1,2,2)$ maps
each latent frame into $30\times52=1560$ visual tokens. Thus, each 5-second clip
contains $21\times1560=32760$ self-attention tokens. Autoregressive generation is
performed in chunks of 3 latent frames.

\subsection{VideoMLA Block}
\label{app:mla_block}

Each temporal self-attention layer follows the notation of Section~3. Given an
attention input $x_t\in\mathbb{R}^d$, VideoMLA first forms a shared content cache
latent
\[
c_t^{KV}=W_{\downarrow}^{KV}x_t\in\mathbb{R}^{d_c},
\]
and a query latent
\[
c_t^Q=W_{\downarrow}^{Q}x_t\in\mathbb{R}^{d_q}.
\]
In the default model, we use
\[
d_c=192,\qquad d_q=768,\qquad
d_h^{\mathrm{nope}}=96,\qquad d_h^{\mathrm{rope}}=32,
\]
with $d_h^{\mathrm{nope}}+d_h^{\mathrm{rope}}=d_h=128$. The down-projection shapes are
\[
W_{\downarrow}^{KV}\in\mathbb{R}^{192\times1536},\qquad
W_{\downarrow}^{Q}\in\mathbb{R}^{768\times1536}.
\]
Both $c_t^{KV}$ and $c_t^Q$ are normalized by RMSNorm before the corresponding
up-projections.

The content key and value for head $h$ are reconstructed from the shared cache
latent:
\[
k^{\mathrm{nope}}_{t,h}=W^K_{\uparrow,h}c_t^{KV},
\qquad
v_{t,h}=W^V_{\uparrow,h}c_t^{KV}.
\]
Aggregating all heads, the projection shapes are
\[
W^K_{\uparrow}\in\mathbb{R}^{(n_h d_h^{\mathrm{nope}})\times d_c}
=\mathbb{R}^{1152\times192},
\]
\[
W^V_{\uparrow}\in\mathbb{R}^{(n_h d_h)\times d_c}
=\mathbb{R}^{1536\times192}.
\]
The NoPE query is reconstructed analogously:
\[
q^{\mathrm{nope}}_{t,h}=W^Q_{\uparrow,h}c_t^Q,
\qquad
W^Q_{\uparrow}\in\mathbb{R}^{(n_h d_h^{\mathrm{nope}})\times d_q}
=\mathbb{R}^{1152\times768}.
\]

The RoPE branch is decoupled from the content cache. For each token, VideoMLA
computes a single head-shared positional key
\[
k_t^R=W_R^K x_t\in\mathbb{R}^{d_h^{\mathrm{rope}}},
\qquad
W_R^K\in\mathbb{R}^{32\times1536},
\]
and per-head positional queries
\[
q_{t,h}^R=W^Q_{R,h}c_t^Q,
\qquad
W_R^Q\in\mathbb{R}^{(n_h d_h^{\mathrm{rope}})\times d_q}
=\mathbb{R}^{384\times768}.
\]
The output projection is the original attention output projection
\[
W_O\in\mathbb{R}^{d\times(n_h d_h)}=\mathbb{R}^{1536\times1536}.
\]

Therefore, each cached token stores only
\[
(c_t^{KV}, k_t^R)\in\mathbb{R}^{d_c+d_h^{\mathrm{rope}}},
\]
rather than dense per-head keys and values. With the default setting, this is
$192+32=224$ scalars per token per layer, compared with
$2n_h d_h=2\cdot12\cdot128=3072$ scalars for dense MHA, corresponding to a
$13.7\times$ cache reduction.

\subsection{NoPE/RoPE Split and 3D RoPE}
\label{app:nope_rope_split}

Each head is split as
\[
d_h=d_h^{\mathrm{nope}}+d_h^{\mathrm{rope}},
\]
where the NoPE subspace is used for content matching and the RoPE subspace is
used for position-aware matching. The per-head query and key are
\[
q_{t,h}=[q^{\mathrm{nope}}_{t,h};q^{\mathrm{rope}}_{t,h}],
\qquad
k_{t,h}=[k^{\mathrm{nope}}_{t,h};k^{\mathrm{rope}}_t],
\]
where
\[
q^{\mathrm{rope}}_{t,h}=\mathrm{RoPE}_{3D}(q^R_{t,h}),
\qquad
k^{\mathrm{rope}}_t=\mathrm{RoPE}_{3D}(k^R_t).
\]
The positional key is shared across heads, while the NoPE keys, NoPE queries, and
values remain head-specific after up-projection.

For the default split \(d_h^{\mathrm{rope}}=32\), the RoPE subspace contains
\(d_h^{\mathrm{rope}}/2=16\) complex frequency pairs. Following the Wan 3D-RoPE
factorization, these pairs are allocated across temporal, height, and width axes as
\[
(6,5,5),
\]
using the highest-frequency bands from the corresponding axis groups.

\subsection{Chunk-Causal Sliding-Window Attention}
\label{app:chunk_causal_attention}

VideoMLA preserves the chunk-causal attention pattern used by the autoregressive
Wan backbone. Tokens within the same 3-latent-frame chunk can attend to one
another, while tokens in a later chunk cannot be attended to by earlier chunks.
For long-horizon generation, attention is restricted to a fixed cache consisting of
one sink latent frame and the most recent six latent frames. Since each latent frame
contains 1560 tokens, the sink occupies 1560 cached token slots and the local window
occupies \(6\times1560\) token slots.

During training, the same chunk-causal and sliding-window structure is enforced
with block-sparse attention masks. During inference, the cache stores
\[
c^{KV}\in\mathbb{R}^{B\times T_{\mathrm{cache}}\times d_c},
\qquad
k^R\in\mathbb{R}^{B\times T_{\mathrm{cache}}\times d_h^{\mathrm{rope}}},
\]
where \(T_{\mathrm{cache}}\) is the number of cached tokens in the sink-plus-window
context. When the cache is full, tokens outside the sink are evicted in FIFO order.
The attention computation reads the active cache, reconstructs the content keys and
values through \(W^K_{\uparrow}\) and \(W^V_{\uparrow}\), applies 3D-RoPE to the
active positional keys, and evaluates the standard per-head attention scores from
Eq.~(7).

\subsection{Long-Horizon RoPE Re-indexing}
\label{app:long_horizon_rope}

For rollouts beyond the 21 latent frames seen during 5-second training, cached
positional keys are stored before RoPE is applied. When an attention window is
assembled, the active cached keys are assigned local temporal coordinates inside
the current sink-plus-window context and are then rotated by \(\mathrm{RoPE}_{3D}\) following \cite{yesiltepe2025infinity}.
The current query chunk is rotated in the same local coordinate system. Thus, both
queries and cached keys use a bounded, window-relative positional frame even after
cache eviction.

This re-indexing keeps the RoPE phase within the short-horizon regime observed by
the backbone during training, while allowing the generated video to extend beyond
the original 21-latent-frame clip length. All reported long-video rollouts use one
sink latent frame, a six-latent-frame local window, and the 4-step student sampler.

\subsection{Training Pipeline}
\label{app:training_pipeline}

We train the same VideoMLA architecture in three stages on 8 NVIDIA B200 GPUs,
using FSDP full sharding, bf16 mixed precision, AdamW with \(\beta_1=0\) and
\(\beta_2=0.999\), and the rectified-flow denoising objective.

\paragraph{Stage 1: Teacher Forcing.}
We initialize the MLA projections from an SVD-style decomposition of the pretrained
Wan dense attention matrices at the target configuration
\[
(d_c,d_q,d_h^{\mathrm{nope}},d_h^{\mathrm{rope}})
=(192,768,96,32).
\]
The model is then trained as a chunk-causal flow-matching student with clean
previous-block context from the teacher-encoded latents. We use learning rate
\(5\times10^{-6}\), per-GPU batch size 1, total batch size 2, gradient checkpointing,
1000 training timesteps, and timestep shift 5.0.

\paragraph{Stage 2: Consistency Distillation.}
Starting from the Stage-1 checkpoint, we distill the model to a 4-step sampling
schedule
\[
[1000,750,500,250].
\]
We use timestep shift 5.0 and classifier-free guidance scale 3.0. The generator
learning rate is \(2\times10^{-6}\), the critic learning rate is \(4\times10^{-7}\),
and the total batch size is 2 with gradient checkpointing enabled.

\paragraph{Stage 3: Distribution Matching Distillation.}
Finally, we initialize from the Stage-2 checkpoint at iteration 2500 and fine-tune
with distribution matching distillation on the same 4-step schedule. The real score
is provided by the frozen teacher and the fake score is learned online. We use five
critic updates per generator update, EMA weight 0.99 starting from step 1, generator
learning rate \(2\times10^{-6}\), critic learning rate \(4\times10^{-7}\), guidance
scale 3.0, and timestep shift 5.0. The total batch size is 16, obtained with per-GPU
batch size 8 on 8 GPUs and gradient accumulation of 2.

The Stage-3 checkpoint is used for all reported VideoMLA results, including the
long-horizon evaluations.

\section{Inference-Time Reparameterization}
\label{sec:appendix_inference_reparam}

The training-time formulation in Section~\ref{sec:method} is written to make the connection to standard multi-head attention explicit: each cached token stores $(c_j^{KV}, k_j^R)$, and the per-head NoPE keys and values are reconstructed through Eq.~\ref{eq:reconstruct} before applying the usual attention computation. This is convenient during training because it allows \method{} to reuse the same block-causal masks, sink-token logic, and attention kernels as the dense baseline. At inference, however, explicitly reconstructing $k_{j,h}^{\mathrm{nope}}$ and $v_{j,h}$ would partially undo the benefit of latent caching by materializing dense per-head tensors after every cache read. We therefore use an equivalent reparameterization that keeps the cache and the attention computation in latent form.

For the score computation, the only contribution of the reconstructed NoPE key is through its inner product with the reconstructed NoPE query. Substituting Eqs.~\ref{eq:cq}, \ref{eq:qnope}, and \ref{eq:reconstruct} into the content term of Eq.~\ref{eq:score} gives
\begin{align}
q_{i,h}^{\mathrm{nope}} \cdot k_{j,h}^{\mathrm{nope}}
&=
\left(W_{\uparrow,h}^{Q} c_i^Q\right)^\top
\left(W_{\uparrow,h}^{K} c_j^{KV}\right) \nonumber \\
&=
\left(c_i^Q\right)^\top
\left(W_{\uparrow,h}^{Q}\right)^\top W_{\uparrow,h}^{K}
c_j^{KV} \nonumber \\
&=
\left(c_i^Q\right)^\top A_h c_j^{KV},
\label{eq:absorb-q}
\end{align}
where
\begin{equation}
A_h =
\left(W_{\uparrow,h}^{Q}\right)^\top W_{\uparrow,h}^{K}
\in \mathbb{R}^{d_q \times d_c}.
\end{equation}
The matrix $A_h$ depends only on learned parameters and is independent of the current sequence, cache contents, diffusion timestep, and rollout position. It can therefore be precomputed once when the model is loaded. During inference, the NoPE content score for head $h$ is computed directly from the query latent $c_i^Q$ and the cached content latent $c_j^{KV}$, without forming either $q_{i,h}^{\mathrm{nope}}$ or $k_{j,h}^{\mathrm{nope}}$ as explicit per-head vectors.

The value path admits an analogous absorption. Let $W_h^O$ denote the slice of the output projection applied to the output of head $h$. Using Eq.~\ref{eq:reconstruct},
\begin{align}
W_h^O v_{j,h}
&=
W_h^O W_{\uparrow,h}^{V} c_j^{KV} \nonumber \\
&=
B_h c_j^{KV},
\label{eq:absorb-v}
\end{align}
where
\begin{equation}
B_h =
W_h^O W_{\uparrow,h}^{V}.
\end{equation}
Thus, the value up-projection can also be folded into the output mixer. In practice, after the attention weights for head $h$ are computed, the weighted sum can be accumulated over the cached latents $c_j^{KV}$ and then projected by $B_h$, rather than first reconstructing all dense values $v_{j,h}$ and then applying the output projection.

The RoPE branch is kept separate from this absorption. The cache stores the unrotated, head-shared positional key $k_j^R$ from Eq.~\ref{eq:krope}. When the active attention window is assembled, $k_j^R$ is rotated by $\mathrm{RoPE}_{\mathrm{3D}}(\cdot)$ using the current window indexing, and the positional score term in Eq.~\ref{eq:score} is computed as
\begin{equation}
q_{i,h}^{\mathrm{rope}} \cdot k_j^{\mathrm{rope}}.
\end{equation}
This separation is important because RoPE is position-dependent and cannot be folded into a fixed parameter matrix in the same way as the NoPE content projections. Storing $k_j^R$ unrotated also preserves the ability to re-index cached tokens within a sliding window, as described in Section~\ref{sec:method:rope}.

After reparameterization, the inference-time cache is never expanded into dense per-head keys and values. Each cached token contributes only a content latent $c_j^{KV}\in\mathbb{R}^{d_c}$ and a head-shared positional key $k_j^R\in\mathbb{R}^{d_h^{\mathrm{rope}}}$. Therefore, the per-token cached state remains
\begin{equation}
d_c + d_h^{\mathrm{rope}},
\end{equation}
instead of the dense baseline cost
\begin{equation}
2n_h d_h.
\end{equation}
For an attention window of size $W$, cache memory traffic is reduced from
\begin{equation}
\mathcal{O}\!\left(W\,2n_h d_h\right)
\end{equation}
to
\begin{equation}
\mathcal{O}\!\left(W\left(d_c+d_h^{\mathrm{rope}}\right)\right).
\end{equation}
With the default configuration $d_c=192$ and $d_h^{\mathrm{rope}}=32$, this corresponds to $224$ cached scalars per token per layer, compared with $2n_hd_h=3072$ scalars for dense MHA. The reparameterization therefore preserves the mathematical attention computation of the training-time formulation while ensuring that the inference-time implementation realizes the intended latent-cache memory and bandwidth savings.

\section{Additional Ablations}

Table~\ref{tab:ablations} studies two architectural choices: the latent KV dimension $d_c$ and the NoPE/RoPE channel split. The latent dimension controls the main quality--efficiency trade-off. At $d_c=64$, VideoMLA gives the largest cache compression and memory headroom, but the budget is too restrictive: both semantic and quality scores drop, consistent with the loss of fine-grained visual details under overly aggressive compression. Increasing to $d_c=128$ largely recovers quality while retaining a large compression ratio. Further increasing to $d_c=256$ or $512$ gives only marginal gains, but substantially reduces the memory advantage. This suggests that the useful operating regime is not the largest possible latent dimension, but the smallest budget that preserves task-relevant video features.

The NoPE/RoPE split also has a clear effect. With only $16$ RoPE channels, positional capacity is too limited, leading to weak temporal and spatial anchoring. Conversely, the RoPE-heavy $32/96$ split leaves too little capacity for cached content and hurts semantic fidelity. The balanced $64/64$ setting improves over these extremes but remains below the content-heavy default. The best result comes from the $96/32$ split, indicating that streaming video benefits from allocating most channels to the cached content path while retaining a smaller dedicated RoPE subspace for positional structure.

\begin{table}[t]
\centering
\vspace{-0.5em}
\small
\setlength{\tabcolsep}{4pt}
\renewcommand{\arraystretch}{1.15}

\begin{minipage}[t]{0.45\linewidth}
\centering
\textbf{(a) Latent dimension $d_c$}
\vspace{0.25em}

\begin{tabular}{c|ccc|cc}
\toprule
\rowcolor{verylightgray}
\boldmath$d_c$ &
\textbf{Semantic}$\uparrow$ &
\textbf{Quality}$\uparrow$ &
\textbf{Total}$\uparrow$ &
\textbf{Mem.}$\downarrow$ &
\textbf{FPS}$\uparrow$ \\
\midrule
64  & 77.42 & 79.18 & 78.30 & 32.00$\times$ & 26.93 \\
128 & 82.16 & 84.31 & 83.24 & 19.20$\times$ & 27.00 \\
256 & 82.74 & 84.58 & 83.66 & 10.67$\times$ & 26.93 \\
512 & 82.41 & 84.39 & 83.40 & 5.65$\times$  & 26.79 \\
\bottomrule
\end{tabular}
\end{minipage}
\hspace{0.08\linewidth}
\begin{minipage}[t]{0.45\linewidth}
\centering
\textbf{(b) NoPE/RoPE split}
\vspace{0.25em}

\begin{tabular}{cc|ccc}
\toprule
\rowcolor{verylightgray}
\boldmath$d_h^{\mathrm{nope}}$ &
\boldmath$d_h^{\mathrm{rope}}$ &
\textbf{Semantic}$\uparrow$ &
\textbf{Quality}$\uparrow$ &
\textbf{Total}$\uparrow$ \\
\midrule
112 & 16 & 74.62 & 78.31 & 76.47 \\
64  & 64 & 79.54 & 82.12 & 80.83 \\
32  & 96 & 75.88 & 80.74 & 78.31 \\
96  & 32 & 83.02 & 84.76 & 83.89 \\
\bottomrule
\end{tabular}
\end{minipage}

\vspace{0.5em}
\caption{\textbf{Ablation studies.} Left: sweep over latent KV dimension $d_c$.
Right: decoupled RoPE dimension ablation with $d_h^{\mathrm{nope}}+d_h^{\mathrm{rope}}=128$.
\textbf{Mem.}: KV cache compression ratio relative to dense per-token KV cache. \textbf{FPS}: throughput at batch size 1 on $1\times$H100 80\,GB.}
\label{tab:ablations}
\vspace{-0.5em}
\end{table}

\end{document}